\pdfoutput=1
\documentclass[11pt]{article}

\usepackage[]{ACL2023}

\usepackage{times}
\usepackage{latexsym}

\usepackage[T1]{fontenc}
\usepackage[utf8]{inputenc}

\usepackage{microtype}

\usepackage{inconsolata}
\definecolor{myblue}{cmyk}{1,0.75,0,0.2}

\newcommand\ie{\textit{i.e.}}
\newcommand\eg{\textit{e.g.}}
\newcommand\etc{\textit{etc}}

\newcommand\wrt{\textit{w.r.t.}}

\usepackage{xcolor}
\usepackage{bm, amsfonts, amsmath}
\usepackage{algorithm, algpseudocode}
\usepackage{newfloat,multirow,booktabs}
\usepackage{amssymb}
\usepackage{graphicx}

\usepackage{arydshln}
\usepackage{subfigure}
\usepackage{pifont}

\title{CoLaDa: A Collaborative Label Denoising Framework \\ 
for Cross-lingual Named Entity Recognition}

\author{Tingting Ma$^{1}$\footnotemark[1] , 
Qianhui Wu$^{2}$, 
Huiqiang Jiang$^{2}$, \\
\bf B\"orje F. Karlsson$^{2}$, Tiejun Zhao$^{1}$\footnotemark[2],
Chin-Yew Lin$^{2}$ \\ 
$^{1}$Harbin Institute of Technology, Harbin, China \quad
$^{2}$Microsoft \\
\tt hittingtingma@gmail.com \\
\tt \{qianhuiwu, hjiang, borjekar, cyl\}@microsoft.com \\
\tt tjzhao@hit.edu.cn 
}

\begin{document}
\maketitle

\renewcommand{\thefootnote}{\fnsymbol{footnote}}
\footnotetext[1]{Work during internship at Microsoft.}
\footnotetext[2]{Corresponding author.}
\renewcommand{\thefootnote}{\arabic{footnote}}

\begin{abstract}
Cross-lingual named entity recognition (NER) aims to train an NER system that generalizes well to a target language by leveraging labeled data in a given source language.
Previous work alleviates the data scarcity problem by translating source-language labeled data or performing knowledge distillation on target-language unlabeled data. However, these methods may suffer from label noise due to the automatic labeling process. In this paper, we propose \textbf{CoLaDa}, a \textbf{Co}llaborative \textbf{La}bel \textbf{D}enoising Fr\textbf{a}mework, to address this problem. Specifically, we first explore a \textit{model-collaboration}-based denoising scheme that enables models trained on different data sources to collaboratively denoise pseudo labels used by each other. We then present an \textit{instance-collaboration}-based strategy that considers the label consistency of each token's neighborhood in the representation space for denoising. Experiments on different benchmark datasets show that the proposed CoLaDa achieves superior results compared to previous methods, especially when generalizing to distant languages.\footnote{Our code is available at \url{https://github.com/microsoft/vert-papers/tree/master/papers/CoLaDa}.}
\end{abstract}

\section{Introduction}
The named entity recognition (NER) task aims to locate and classify entity spans in a given text into predefined entity types.
It is widely used for many downstream applications, such as relation extraction and question answering.
Deep neural networks have made significant progress on this task leveraging large-scale human-annotated data for training.
However, fine-grained token-level annotation makes it costly to collect enough high-quality labeled data, especially for low-resource languages.
Such scenarios motivate the research on \textit{zero-shot} cross-lingual NER, which attempts to leverage labeled data in a rich-resource source language to solve the NER task in a target language without annotated data. 

\begin{figure}[t]
    \centering
    \includegraphics[width=\columnwidth]{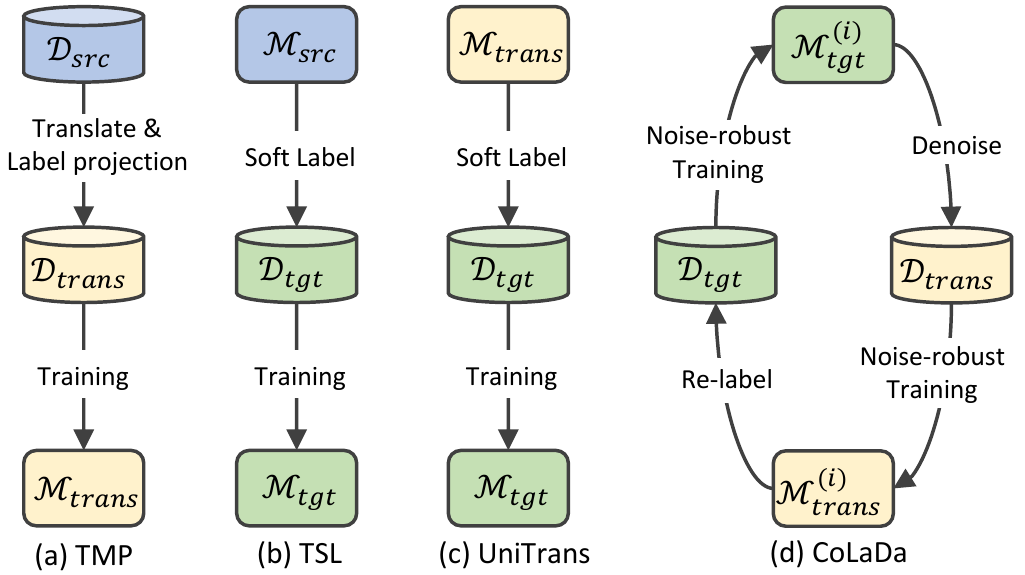}
    \caption{Comparison between previous methods (\textbf{a/b/c}) and our CoLaDa at the $i$-th iteration (\textbf{d}) CoLaDa starts at $\mathcal{M}_{tgt}^0$ and performs denoising iteratively.
    $\mathcal{D}_{src}$: Source-language labeled data. $\mathcal{D}_{trans}$: Translation data.
    $\mathcal{D}_{tgt}$: Target-language unlabeled data with pseudo-labels generated by NER models.
    $\mathcal{M}_{src/trans/tgt}$: NER model learned on $\mathcal{D}_{src/trans/tgt}$.}
    \label{fig:model-intro}
\end{figure}

Recent attempts at cross-lingual NER can be roughly categorized from two aspects: learning language-independent features via feature alignment~\citep{huang-2019-crossadv, keung-2019-adversarial} and learning language-specific features from automatically labeled target-language data~\citep{wu-2020-smts, wu-2020-unitrans}. 
Despite bringing great success to cross-lingual NER, the former line of research misses exploiting language-specific features and thus shows substandard performance, especially when transferring to distant languages, \eg{,} from English to Arabic~\citep{fu-2023-heter}. 
Hence, a series of studies focuses on the latter category, which typically creates pseudo-labeled target-language data and uses it to perform conventional supervised learning or teacher-student learning.
For example, as shown in Fig~\ref{fig:model-intro}(a),
earlier studies \citep{ehrmann-2011-building,mayhew-2017-cheap,xie-2018-word,jain-2019-entproj}, such as TMP \cite{jain-2019-entproj}, first translate labeled data in the source language and then perform label projection. 
Recently, several approaches have utilized a weak model, which could be an NER model either trained on the source language's labeled data as in TSL \cite{wu-2020-smts}, or further finetuned on the generated translation data as in UniTrans \cite{wu-2020-unitrans}, to annotate the unlabeled target-language data for improvement,
as shown in Fig~\ref{fig:model-intro}(b) and Fig~\ref{fig:model-intro}(c).

Unfortunately, these methods inevitably suffer from the label noise induced by inaccurate translation and label projection, or the weak model's limited capability.
Although some methods are proposed to mitigate the label noise problem by additionally training an instance selector~\citep{liang-2021-rikd, chen-2021-advpicker} or designing heuristic rules for data selection~\citep{ni-2017-weakly}, they independently manipulate either the translation data ($\mathcal{D}_{trans}$)~\citep{ni-2017-weakly} or the target-language data ($\mathcal{D}_{tgt}$) pseudo-labeled by NER models trained in the source language~\citep{liang-2021-rikd, chen-2021-advpicker}.
Hence, all these methods ignore the complementary characteristics between both for denoising.
Particularly, from the \textit{text view}, $\mathcal{D}_{tgt}$ is collected from a natural text distribution of the target-language data,
while $\mathcal{D}_{trans}$ can be regarded as a way of data augmentation to provide more lexicon variants. 
From the \textit{labeling function view}, labels of $\mathcal{D}_{trans}$ are obtained via the label projection algorithm, which have little association with those of $\mathcal{D}_{tgt}$ generated by NER models.

With such consideration, we propose \textbf{a \textit{model-collaboration}-based denoising scheme}, which incorporates models trained on both data sources to mutually denoise the pseudo-labels of both data sources in an iterative way.
As shown in Fig~\ref{fig:model-intro}(d), we first leverage $\mathcal{M}_{tgt}$ trained on the pseudo-labeled target-language data $\mathcal{D}_{tgt}$ to denoise the translation data annotated by label projection.
In this way, the learned model $\mathcal{M}_{trans}$ will be less affected by noise in the translation data. 
We then employ the improved $\mathcal{M}_{trans}$ to re-label the target-language unlabeled data $\mathcal{D}_{tgt}$.
It is expected that there is less noise in the relabeled data, and thus we can produce a more powerful $\mathcal{M}_{tgt}$.
We perform this procedure for several iterations, so that all the involved data sources and models can be improved in an upward spiral.

\begin{figure}[t]
    \centering
    \includegraphics[width=0.7\columnwidth]{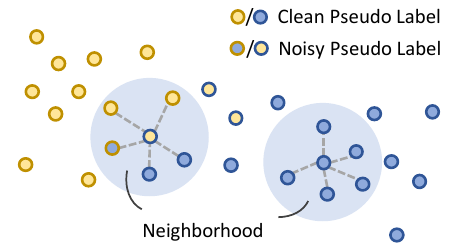}
    \caption{Illustration of the instance collaboration for denoising. Different colors depict different entity types.}
    \label{fig:knn-intro}
\end{figure}
 Moreover, borrowing the idea from anomaly detection~\citep{gu2019statistical} that a given data point's neighborhood information can be used to measure its anomalism, here we find that the similar tokens in the feature space can also collaborate for denoising. 
 Previous studies~\citep{zhai2018classification, xu-2020-deepood} have shown that instances with the same label are more likely to locate close to each other in the representation space.
 Our intuition is that, if a token's label conflicts a lot with labels of other tokens in its neighborhood, then this label is probably noisy.
Therefore, we further propose \textbf{an \textit{instance-collaboration}-based denoising strategy} to explore the neighborhood structure of each token for denoising, as shown in Figure~\ref{fig:knn-intro}.
Specifically, we utilize the label consistency of each token's neighborhood in the representation space to re-weight the soft-labeled examples in knowledge distillation.

We integrate the instance-collaboration-based denoising strategy into the model-collaboration-based denoising scheme and propose a \textbf{Co}llaborative \textbf{La}bel \textbf{D}enoising fr\textbf{a}mework, \ie,  \textbf{CoLaDa}, for cross-lingual NER.
We conduct extensive experiments on two popular benchmarks covering six languages for evaluation.
Experimental results show that our method outperforms existing state-of-the-art methods.
Qualitative and quantitative analyses further demonstrate the effectiveness of our framework in reducing the data noise.

\section{Problem Formulation}

Here we take the typical sequence labeling formulation for the named entity recognition task. 
Given a sequence with $L$ tokens $\bm{x}=(x_1,\dots, x_L)$ as the input text, an NER system is excepted to assign each token $x_i$ with a label $y_i$. 

In this paper, we assume to have the labeled training data $\mathcal{D}_{src}=\{(\bm{x}^s, \bm{y}^s)\}$ in the source language, the unlabeled data $\mathcal{D}_{tgt}=\{\bm{x}^{u}\}$ from the target language, 
and translation data $\mathcal{D}_{trans}=\{(\bm{x}^{t}, \bm{y}^{t})\}$ obtained by data projection from $\mathcal{D}_{src}$. 
Our goal is to train an NER model $\mathcal{M}$ that can generalize well to the target language utilizing these resources.

\section{CoLaDa Framework}

\begin{figure*}[htbp]
    \centering
    \includegraphics[width=1.85\columnwidth,height=0.85\columnwidth]{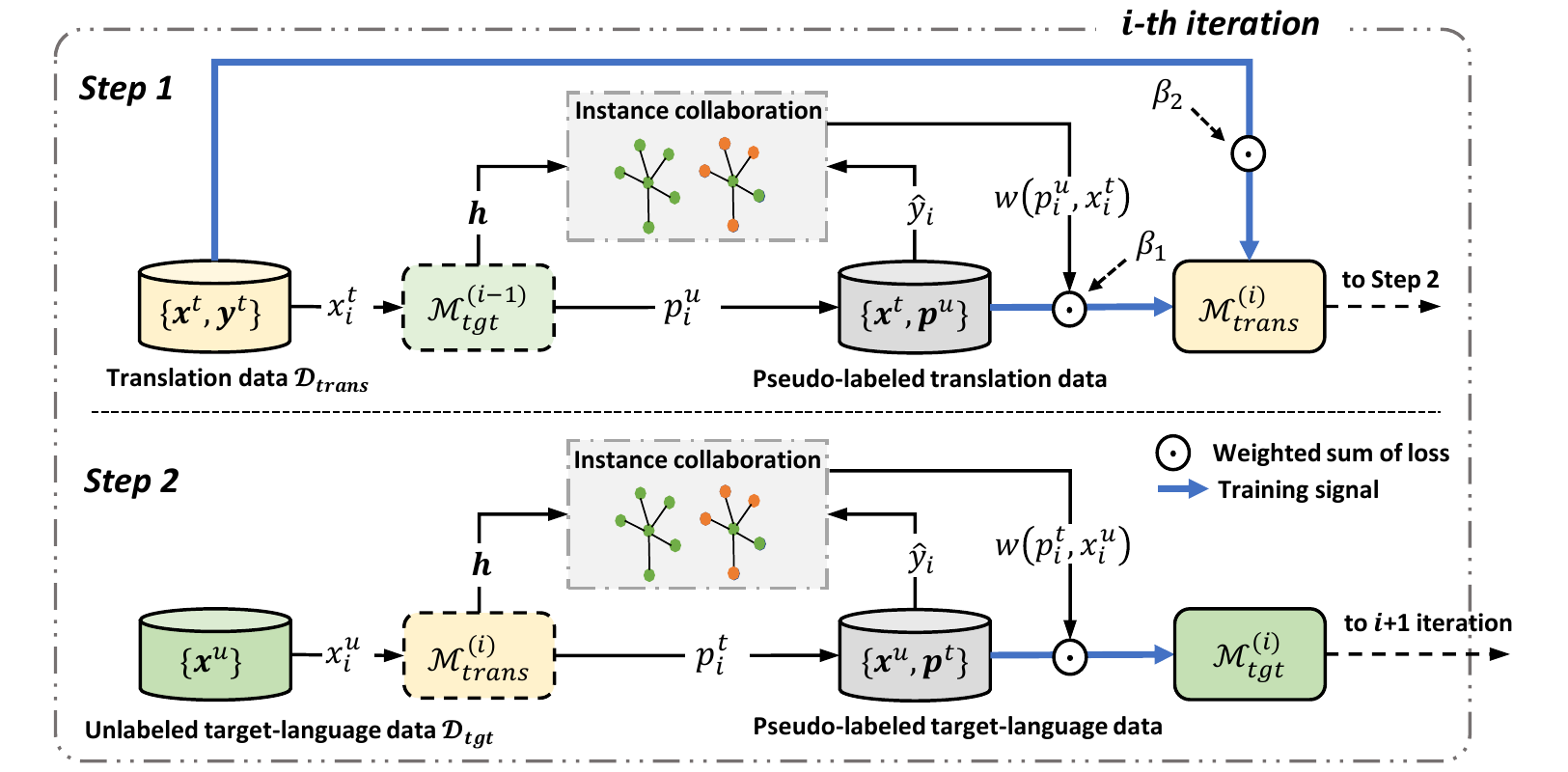}
    \caption{Framework of \textit{CoLaDa}, which is an \textbf{iterative model-collaboration} process with two steps: 1) Step 1: noise-robust training on translation data with the collaborator $\mathcal{M}_{tgt}^{(i-1)}$, 2) Step 2: noise-robust training on unlabeled target-language data with the collaborator $\mathcal{M}_{trans}^{(i)}$. The \textbf{instance-collaboration} is used to re-weight the noisy labels from a teacher model in both steps. $\mathcal{M}^{(i)}_{trans}/\mathcal{M}^{(i)}_{tgt}$: model trained on $\mathcal{D}_{trans}/\mathcal{D}_{tgt}$ at $i$-th iteration.}
    \label{fig:colada}
\end{figure*}

Figure~\ref{fig:colada} depicts an overview of the CoLaDa framework.
It is an iterative model-collaboration-based denoising framework which consists of two steps: noise-robust learning on translation data and noise-robust learning on unlabeled target-language data.
An instance-collaboration-based denoising strategy (Sec~\ref{sec:knn}) is then integrated into the model-collaboration-based denoising procedure (Sec~\ref{sec:model_collaboration}).

\subsection{Instance Collaboration for Denoising}
\label{sec:knn}
Previous work~\citep{zhai2018classification, xu-2020-deepood} indicates that tokens with the same labels are more likely to locate close to each other in the representation space of a deep neural network.
If the label of a given token is inconsistent with lots of its neighbors, this token would be isolated from other tokens with the same label in the feature space, and hence its label  is more likely to be noisy.
Therefore, we propose instance-collaboration-based denoising, which evaluates the reliability of a given token's label by measuring the label consistency of its neighborhood, and then uses the reliability score to  weight the noisy labels from a teacher model $\mathcal{M}$ for knowledge distillation on data $\mathcal{D} = \{\bm{x}\}$.
Noisy labels are expected to have lower weights than clean ones. 

\paragraph{Create a memory bank.}
We leverage the feature extractor $\mathcal{F}$ of the NER model $\mathcal{M}$ to obtain the hidden representations $\bm{h}=\{h_i\}_{i=1}^{L}$ of each sentence $\bm{x}=\{x_i\}_{i=1}^L\in \mathcal{D}$:

\begin{equation}
    \bm{h}={\mathcal{F}(\bm{x})}.
\end{equation}
We then construct a memory bank $\mathcal{B}_{\mathcal{D}}=\{h\}$ to store the hidden representations of all tokens in $\mathcal{D}$. 

\paragraph{Compute label consistency.}
Given a token $x_i$,
we retrieve its $K$-nearest neighbors $\mathcal{N}_K(x_i)$ in $\mathcal{B}_{\mathcal{D}}$ using cosine similarity.
Let $p_i$ denote the soft label (\ie, the probability distribution over the entity label set) assigned by the teacher model $\mathcal{M}$ for $x_i$.
We measure the label consistency of $x_i$, \ie, $\lambda(p_i; x_i)$, by calculating the fraction of $x_i$'s neighbors that are assigned with the same labels as $x_i$ in $\mathcal{N}_K(x_i)$:
\begin{equation}
\lambda(p_i; x_i) = \frac{1}{K}\sum_{x_j \in \mathcal{N}_k(x_i)} I(\hat{y}_j = \hat{y}_i),
\label{eq:knn-frac}
\end{equation}
where
$\hat{y}_i = \arg{}\max{(p_i)}$ is the pseudo entity label corresponding to the maximum probability in $p_i$.
Similarly, $\hat{y}_j$ is the pseudo entity label corresponding to $x_j$.
$I$ is the indicator function.

\paragraph{Produce a reliability score.}
We use the label consistency $\lambda(p_i; x_i)$ to compute the reliability score of the soft label $p_i$, which is further used as the weight of $p_i$ during model learning (see \ref{sec:model_collaboration}).
Considering that different entity types may contain different levels of label noise and show different statistics on label consistency, 
here we present a class-adaptive reliability score for weighting:
\begin{gather}
w(p_i; x_i) = \textrm{Sigmoid}\left(\alpha\left(\lambda(p_i; x_i) - \mu(\hat{y}_i)\right)\right),
\label{eq:knn}
\end{gather}
% \end{equation}
where $\mu(\hat{y}_i)$ denote the mean of all $\lambda(p_j; x_j)$ where $\arg{}\max{(p_j)} = \hat{y}_i$ and $x_j\in\mathcal{D}$.
$\alpha > 0$ is a hyper-parameter that controls the sharpness of the weighting strategy. 
If $\alpha \to 0$, all tokens have equal weights.
If $\alpha \to \infty$,
tokens whose label consistency is larger than the average label consistency \wrt{} its pseudo label will be weighted with 1 and those with smaller consistency will be dropped.

\subsection{Model Collaboration for Denoising}
Here we elaborate on the details of the two noise-robust training processes.
Algorithm~\ref{code} depicts the overall training procedure of CoLaDa.

\label{sec:model_collaboration}
\paragraph{Noise-robust training on translation data.}
Assuming the availability of a collaborator $\mathcal{M}_{tgt}$\footnote{For the first iteration, we use an NER model trained on the source language labeled data $\mathcal{D}_{src}$. For the later iterations ($i>1$), we use the model from the noise-robust-training on target-language unlabeled data in the previous iteration ($i-1$).}
trained on pseudo-labeled target-language data $\mathcal{D}_{tgt}$,
here we focus on leveraging $\mathcal{M}_{tgt}$ to reduce the label noise in the translation data $\mathcal{D}_{trans} = \{(\bm{x}^t, \bm{y}^t)\}$, with which we further deliver a more powerful model $\mathcal{M}_{trans}$.

Specifically, given a sentence $(\bm{x}^t, \bm{y}^t)\in\mathcal{D}_{trans}$,
we first obtain the soft label $p_i^u$ of each $x_i^t \in \bm{x}^t$ from the collaborator $\mathcal{M}_{tgt}$.
Then, we take both the one hot label $y_i^t$ and the soft label $p_i^u$ as the supervision to train the model $\mathcal{M}_{trans}$.\footnote{The student model $\mathcal{M}_{trans}$ is initialized from $\mathcal{M}_{tgt}$ to equip the knowledge of real target-language text distribution for better generalization during test.}
Denote the output probability distribution of $\mathcal{M}_{trans}$ for $x_i^t$ as $\hat{p}_i^t$.
The loss function \wrt{} $\bm{x}^t$ is defined as:
\begin{equation}
    \mathcal{L}^{\bm{x}^t}=\frac{1}{L}\sum_{i=1}^L \left(\beta_1 \textrm{CE}(\hat{p}_i^t, p_i^u)+\beta_2\textrm{CE}(\hat{p}_i^t, y_i^{t})\right),
    \label{eq:step1}
\end{equation}
where $\textrm{CE}(\cdot, \cdot)$ denotes the cross-entropy loss, $L$ is the sentence length, $\beta_1$ and $\beta_2$ are weighting scalars.
Here we further incorporate the instance-collaboration-based denoising strategy (\ref{sec:knn}) to provide a token-level reliability evaluation to the supervision from the collaborator $\mathcal{M}_{tgt}$ via:
\begin{equation}
\beta_1(x_i^t) \leftarrow \beta_1 * w(p_i^u, x_i^t),
\label{eq:beta}
\end{equation}
where $w(p_i^u, x_i^t)$ is calculated by Eq.~(\ref{eq:knn}).

\paragraph{Noise-robust training on target-language unlabeled data.}
Here we leverage $\mathcal{M}_{trans}$ obtained via the above noise-robust training on translation data to provide high-quality supervision for $\mathcal{D}_{tgt}=\{\bm{x}^u\}$.
By performing knowledge distillation on $\mathcal{D}_{tgt}$, the student model $\mathcal{M}_{tgt}$ is supposed to benefit from the unlabeled data drawn from the real text distribution in the target language with the knowledge from the teacher model $\mathcal{M}_{trans}$.

Specifically, given a sentence $\bm{x}^u \in \mathcal{D}_{tgt}$, we first
utilize $\mathcal{M}_{trans}$ to predict soft label $p_i^t$ for each token $x_i^u\in \bm{x}^u$.
Then, we integrate the instance-collaboration-based denoising technique into the learning process.
The loss function \wrt{} $\bm{x}^u$ to train the student model $\mathcal{M}_{tgt}$ can be formulated as:
\begin{equation}
    \mathcal{L}^{\bm{x}^u} = \frac{1}{L}\sum_{i=1}^L w(p_i^t, x_i^u) \cdot \textrm{CE}(\hat{p}_i^u, p_i^t),
\label{eq:step2}
\end{equation}
where $\hat{p}_i^u$ denotes the output probability distribution of $\mathcal{M}_{tgt}$ for the $i$-th token $x_i^u$ and $w(p_i^t, x_i^u)$ is calculated by Eq.~(\ref{eq:knn}).

\begin{algorithm}[t]
    \small
	\caption{Pseudo code of CoLaDa.} 
    \textbf{Input}: an NER model $\mathcal{M}_{src}$ trained on $\mathcal{D}_{src}$, translation data $\mathcal{D}_{trans}$, the unlabeled data $\mathcal{D}_{tgt}$, the maximum iteration T. 
	\begin{algorithmic}[1]
         \State $\mathcal{M}_{tgt}^{(0)} \leftarrow 
         \mathcal{M}_{src}$ \Comment{Initialization}
    	\For {$i=1,2,\ldots$, T} 
        \State \textit{\# \textcolor{blue}{Step 1}: Noise-robust training on} $D_{trans}$
            \State 
            Inference $\mathcal{M}^{(i-1)}_{tgt}$ on $\mathcal{D}_{trans}=\{(\bm{x}^t, \bm{y}^{t})\}$ to get the predictions $\widehat{\mathcal{D}}_{trans}=\{(\bm{x}^t, \bm{p}^u)\}$
            \State 
            Get $\bm{w}$ for $(\bm{x}^t, \bm{p}^u)\in \widehat{\mathcal{D}}_{trans}$ with $\mathcal{M}_{tgt}^{(i-1)}$, Eq.(\ref{eq:knn})
            \State
            Train $\mathcal{M}_{trans}^{(i)}$ with loss on $(\bm{x}^t, \bm{y}^t, \bm{p}^u, \bm{w})$, Eq.(\ref{eq:step1})
        \State \textit{\# \textcolor{blue}{Step 2}: Noise-robust training on} $D_{tgt}$
        \State
            Inference $\mathcal{M}^{(i)}_{trans}$ on $\mathcal{D}_{tgt}=\{\bm{x}^u\}$ to get the predictions $\widehat{\mathcal{D}}_{tgt}=\{(\bm{x}^u, \bm{p}^t)\}$
            \State 
            Get $\bm{w}'$ for $(\bm{x}^u, \bm{p}^t)\in \widehat{D}_{tgt}$ with $\mathcal{M}_{trans}^{(i)}$, Eq.(\ref{eq:knn})
            \State
            Train $\mathcal{M}_{tgt}^{(i)}$ with loss on $(\bm{x}^u, \bm{p}^t, \bm{w}')$, Eq.(\ref{eq:step2})       
		\EndFor
	\end{algorithmic} 
    \textbf{Output}: an NER model $\mathcal{M}_{tgt}^{(T)}$. 
    \label{code}
\end{algorithm}

\section{Experiments}

\subsection{Experiment Settings}
\paragraph{Datasets}
We conduct experiments on two standard cross-lingual NER benchmarks: CoNLL~\citep{tjong-2002-conll,tjong-2003-conll} and WikiAnn~\citep{pan-2017-wikiann}. CoNLL contains four languages: English (en) and German (de) from the CoNLL-2003\footnote{https://www.clips.uantwerpen.be/conll2003/ner/} NER shared task~\citep{tjong-2003-conll}, and Spanish (es) and Dutch (nl) from  the CoNLL-2002\footnote{https://www.clips.uantwerpen.be/conll2002/ner/} NER shared task~\citep{tjong-2002-conll}.
This dataset is annotated with four entity types: PER, LOC, ORG, and MISC.
WikiAnn contains an English dataset and datasets in three non-western languages: Arabic (ar), Hindi (hi), and Chinese (zh). Each dataset is annotated with 3 entity types: PER, LOC, and ORG. 
All datasets are annotated with the BIO tagging scheme.
We use the train, development, and test splits as previous work~\citep{wu-2019-suprise, wu-2020-unitrans}.
% Table~\ref{tab:data} in the Appendix shows the dataset statistics.

 We take English as the source language and other languages as the target language, respectively.
 We remove the labels of the training data for the target language and take it as the unlabeled target language data.
 For the CoNLL benchmark, we use the word-to-word translation data provided in UniTrans~\citep{wu-2020-unitrans} for a fair comparison.
 For the WikiAnn benchmark,
 we translate the source data to the target language with the public M2M100~\citep{fan-2020-m2m} translation system and conduct label projection with the marker-based alignment algorithm as \citet{yang-2022-crop}.

 \paragraph{Evaluation} The entity-level micro-F1 on test set of the target language is used as the evaluation metric.
 We report the mean value of 5 runs with different seeds for all the experiments.

\paragraph{Implementation Details} For the base NER model, we stack a linear classifier with softmax over a base encoder such as mBERT. 
We implement our framework with Pytorch 1.7.1\footnote{\url{https://pytorch.org/}}, the \textit{HuggingFace} transformer library~\citep{wolf-2020-transformers}, and use FAISS~\cite{johnson2019billion} for embedding retrieval.
Following \citet{wu-2019-suprise} and \citet{zhou-2022-conner}, we use the multilingual BERT base model~\citep{devlin-2019-bert} and XLM-R~\citep{conneau-2020-xlmr} large model as our base encoders.
Most of our hyper-parameters are set following \citet{wu-2020-unitrans}.
We use AdamW~\citep{loshchilov2018decoupled} as optimizer and train the model on source NER data with the learning rate of 5e-5 for 3 epochs.
The dropout rate is 0.1.
For teacher-student learning, we train the model with a learning rate of 2e-5 for 10 epochs.
% train the student model
We freeze the bottom three layers as \citet{wu-2019-suprise}.
Following \citet{keung-2019-adversarial}, we choose other hyper-parameters according to the target language dev set.
We set K in Eq. (\ref{eq:knn-frac}) to 500 and $\alpha$ in Eq. (\ref{eq:knn}) to 6.
For the first iteration, we start with an NER model trained on the source-language data to denoise the translation data with $\beta_1$ and $\beta_2$ in Eq. (\ref{eq:beta}) setting to 0.5. 
For the following iterations, $\beta_1$ is set to 0.9 and $\beta_2$ is set to 0.1. 
The maximum number of iterations is 8.

\begin{table}[t]
    % \begin{minipage}{\columnwidth}        
    \centering
    \setlength{\tabcolsep}{1.5mm}
    \resizebox{\columnwidth}{!}{
    \begin{tabular}{lcccc}
    \toprule
      \textbf{Method}  & \textbf{de} & \textbf{es} & \textbf{nl} & \textbf{avg} \\
    \midrule
    \midrule
    \textit{mBERT based methods:} &&&&\\
    \midrule
      % TMP~\citep{jain-2019-entproj} & 61.50 & 73.50 & 69.90 & 68.30 \\
      mBERT~\citep{wu-2019-suprise} & 69.56 & 74.96 &  77.57 & 73.57 \\
      AdvCE~\citep{keung-2019-adversarial} & 71.90 & 74.3 & 77.60 & 74.60 \\
      TSL~\citep{wu-2020-smts} & 73.16 & 76.75 & 80.44 & 76.78 \\
      UniTrans~\citep{wu-2020-unitrans} & 74.82 & 79.31 & 82.90 & 79.01 \\
      TOF~\citep{zhang-2021-tof}  & 76.57 & 80.35 & 82.79 & 79.90 \\
      AdvPicker~\citep{chen-2021-advpicker} & 75.01 & 79.00 & 82.90 & 78.97 \\
      RIKD~\citep{liang-2021-rikd} & 75.48 & 77.84 & 82.46 & 78.59 \\
      MTMT~\citep{li-2022-mtmt} & 76.80 & \textbf{81.82} & 83.41 & 80.67 \\
    \midrule
     \textbf{CoLaDa (ours)}  & \textbf{77.30} & 80.43 & \textbf{85.09} & \textbf{80.94} \\
    \midrule
    \midrule
    \textit{XLM-R based methods:} &&&&\\
    \midrule
    % RIKD~\citep{liang-2021-rikd} (base) & 78.40 & 79.46 & 81.40 & avg \\
    MulDA~\citep{liu-2021-mulda} & 74.55 & 78.14 & 80.22 & 77.64 \\
    xTune~\citep{zheng-2021-xtune} & 74.78 & 80.03 & 81.76 & 78.85 \\
    ConNER~\citep{zhou-2022-conner} & 77.14 & 80.50 & 83.23 & 80.29 \\
    \midrule
    \textbf{CoLaDa (ours)} & \textbf{81.12} & \textbf{82.70} & \textbf{85.15} & \textbf{82.99} \\
    \bottomrule
    \end{tabular}
    }
    \caption{F1 scores on CoNLL.}
    \label{tab:conll-res}
\end{table}
% \end{minipage}
% \hfill
% \begin{minipage}{\columnwidth}
\begin{table}[t]
    \centering    \setlength{\tabcolsep}{1.5mm}
    \resizebox{\columnwidth}{!}{
    \begin{tabular}{lcccc}
    \toprule
      \textbf{Method}  & \textbf{ar} & \textbf{hi} & \textbf{zh} & \textbf{avg} \\
    \midrule
    \midrule
    \textit{mBERT based methods:} &&&&\\
    \midrule
      BERT-align~\citep{wu-dredze-2020-explicit} & 42.30 & 67.60 &  52.90 & 54.26 \\
      TSL~\citep{wu-2020-smts} & 43.12 & 69.54 &  48.12 & 53.59 \\
      RIKD~\citep{liang-2021-rikd} & 45.96 & 70.28 & 50.40 & 55.55 \\
      MTMT~\citep{li-2022-mtmt} & 52.77 & 70.76 &  52.26 & 58.59 \\
      % Unitrans(MUSE)~\citep{wu-2020-unitrans} & 43.57 & 66.63 & 51.72 & 53.97 \\
      UniTrans$^\dag$~\citep{wu-2020-unitrans} & 42.90  & 68.76 & 56.08 & 55.91 \\
    \midrule
    \textbf{CoLaDa (ours)} & \textbf{54.26} & \textbf{72.42} & \textbf{60.77} & \textbf{62.48} \\
    \midrule
    \midrule
    \textit{XLM-R based methods:} &&&&\\
    \midrule
    XLM-R~\citep{conneau-2020-xlmr} & 50.84 & 72.17 & 39.23 & 54.08 \\
    % RIKD~\citep{liang-2021-rikd} & 54.46 & 74.42 & 37.48 & avg \\
    ConNER~\citep{zhou-2022-conner} & 59.62 & 74.49 & 39.17 & 57.76 \\
    \midrule
    \textbf{CoLaDa (ours)} & \textbf{66.94} & \textbf{76.69} & \textbf{60.08} & \textbf{67.90} \\
    \bottomrule
    \end{tabular}
    }
    \caption{F1 scores on WikiAnn. $^{\dag}$ denotes results obtained by running their public code on our data.}
    \label{tab:wiki-res}
% \end{minipage}
\end{table}

\subsection{Main Results}
\paragraph{Baselines} We compare our method to previous start-of-the-art baselines as follows: i) feature alignment based methods: mBERT~\citep{wu-2019-suprise}, XLM-R~\citep{conneau-2020-xlmr}, BERT-align~\citep{wu-dredze-2020-explicit}, AdvCE~\citep{keung-2019-adversarial}, and AdvPicker~\citep{chen-2021-advpicker}; ii) translation based methods: MulDA~\citep{liu-2021-mulda}, UniTrans~\citep{wu-2020-unitrans}, and TOF~\citep{zhang-2021-tof}); iii) knowledge distillation based methods: TSL~\citep{wu-2020-smts}, RIKD~\citep{liang-2021-rikd}, and MTMT~\citep{li-2022-mtmt}; iv) consistency based methods: xTune~\citep{zheng-2021-xtune} and ConNER~\citep{zhou-2022-conner}.

\begin{table*}[t]
    \centering
    \small
    \begin{tabular}{lcccccc}
    \toprule
       \textbf{Method}  & \textbf{de} & \textbf{es} & \textbf{nl} & \textbf{ar} & \textbf{hi} & \textbf{zh} \\
    \midrule
       CoLaDa  & \textbf{77.30} & \textbf{80.43} & \textbf{85.09} & \textbf{54.26} & \textbf{72.42} & \textbf{60.77} \\
       \midrule
       1) CoLaDa w/o instance collaboration & 76.08 & 79.94 & 83.86 & 50.98 & 71.31 & 59.64 \\
       \specialrule{0em}{1pt}{1pt}
       \hdashline
       \specialrule{0em}{1pt}{1pt}
       2) CoLaDa w/o translation data denoise & 76.17 & 79.22 & 83.10 & 41.41 & 71.10 & 55.04 \\
       3) CoLaDa w/o iteratively denoise & 75.77 & 79.64 & 83.50 & 47.82 & 71.31 & 57.64 \\
       4) CoLaDa w/o model collaboration & 75.64 & 78.99 & 82.98 & 46.51 & 71.09 & 55.25 \\
       \specialrule{0em}{1pt}{1pt}
       \hdashline
       \specialrule{0em}{1pt}{1pt}
       5) CoLaDa w/o instance \& model collaboration & 74.54 & 79.94 & 82.97 & 42.33 & 70.39 & 55.55 \\
    \bottomrule
    \end{tabular}
    \caption{Ablation study on CoNLL and WikiAnn.}
    \label{tab:ablation}
\end{table*}

\paragraph{Performance Comparison} Tables~\ref{tab:conll-res} and~\ref{tab:wiki-res} show the performance comparison of the proposed CoLaDa and prior start-of-the-art baselines on CoNLL and Wikiann, respectively.
It can be seen that CoLaDa outperforms prior methods with both encoders, achieving a significant improvement of 2.70 F1 scores on average for CoNLL and 10.14 F1 scores on average for WikiAnn with XLM-R as the encoder.
This well demonstrates the effectiveness of our approach. 
Interestingly, CoLaDa shows more significant superiority when transferring to distant target languages in WikiAnn.
The knowledge distillation based baselines (\ie{,} TSL, RIKD, MTMT) struggle on distant languages such as Chinese (zh) due to the noisy predictions from the weak teacher model $\mathcal{M}_{src}$ trained in the source language.
UniTrans, which is developed with the same data sources as ours, shows poor performance, especially in distant languages such as Arabic (ar).
We conjecture that the problem of label noise is even more critical in these distant languages.
Our CoLaDa can better handle noise in both translation data and unlabeled target-language data, thus leading to significant performance gains.

\section{Analysis}

\subsection{Ablation Study}
To further validate the effectiveness of each mechanism in the proposed framework, we introduce the following variants of CoLaDa in an ablation study:
1) \textit{CoLaDa w/o instance collaboration}, where we directly set the reliability score in Eq. (\ref{eq:knn}) to $1$ for all tokens.
2) \textit{CoLaDa w/o translation data denoise}, where we set $\beta_1$ in Eq. (\ref{eq:step1}) to 0.
3) \textit{CoLaDa w/o iteratively denoise}, where we remove the iterative enhancement and only conduct the denoising process for one iteration.
4) \textit{CoLaDa w/o model collaboration}, where we set $\beta_1$ in Eq. (\ref{eq:step1}) to 0, remove the iteration mechanism, and 
directly take the model finetuned on $\mathcal{D}_{trans}$ as the teacher model to train a student model with instance-collaboration-based denoising on $\mathcal{D}_{tgt}$.
5) \textit{CoLaDa w/o instance \& model collaboration}, which further drops the instance-collaboration-based denoising from 4).

Table~\ref{tab:ablation} shows the ablation results. We can draw some in-depth conclusions as follows.

1) \textit{CoLaDa} outperforms \textit{CoLaDa w/o instance collaboration}, 
which highlights the effectiveness of leveraging neighborhood information to reduce label noise in knowledge distillation.

2) \textit{CoLaDa} outperforms \textit{CoLaDa w/o translation data denoise},
which emphasizes the importance of using the collaborator $\mathcal{M}_{tgt}$ to refine labels of translation data,
especially in distant languages where the translation data is noisier (\eg{,} 12.8 F1 drop on Arabic and 5.7 F1 drop on Chinese).

3) \textit{CoLaDa} outperforms \textit{CoLaDa w/o iteratively denoise}, which indicates the necessity of iterative learning: models obtained from the previous iteration should be re-used as the collaborator to further improve label quality in the next iteration.

4) \textit{CoLaDa w/o instance \& model collaboration}, which eliminates all denoising strategies from \textit{CoLaDa}, leads to a significant performance drop, demonstrating the essentiality of label denoising for cross-lingual NER.

\subsection{Analysis of Model Collaboration}
Here we attempt to understand how the two models, \ie, $\mathcal{M}_{trans}$ and $\mathcal{M}_{tgt}$, collaboratively improve each other.
\begin{figure}[h]
    \centering
    \includegraphics[width=0.8\columnwidth]{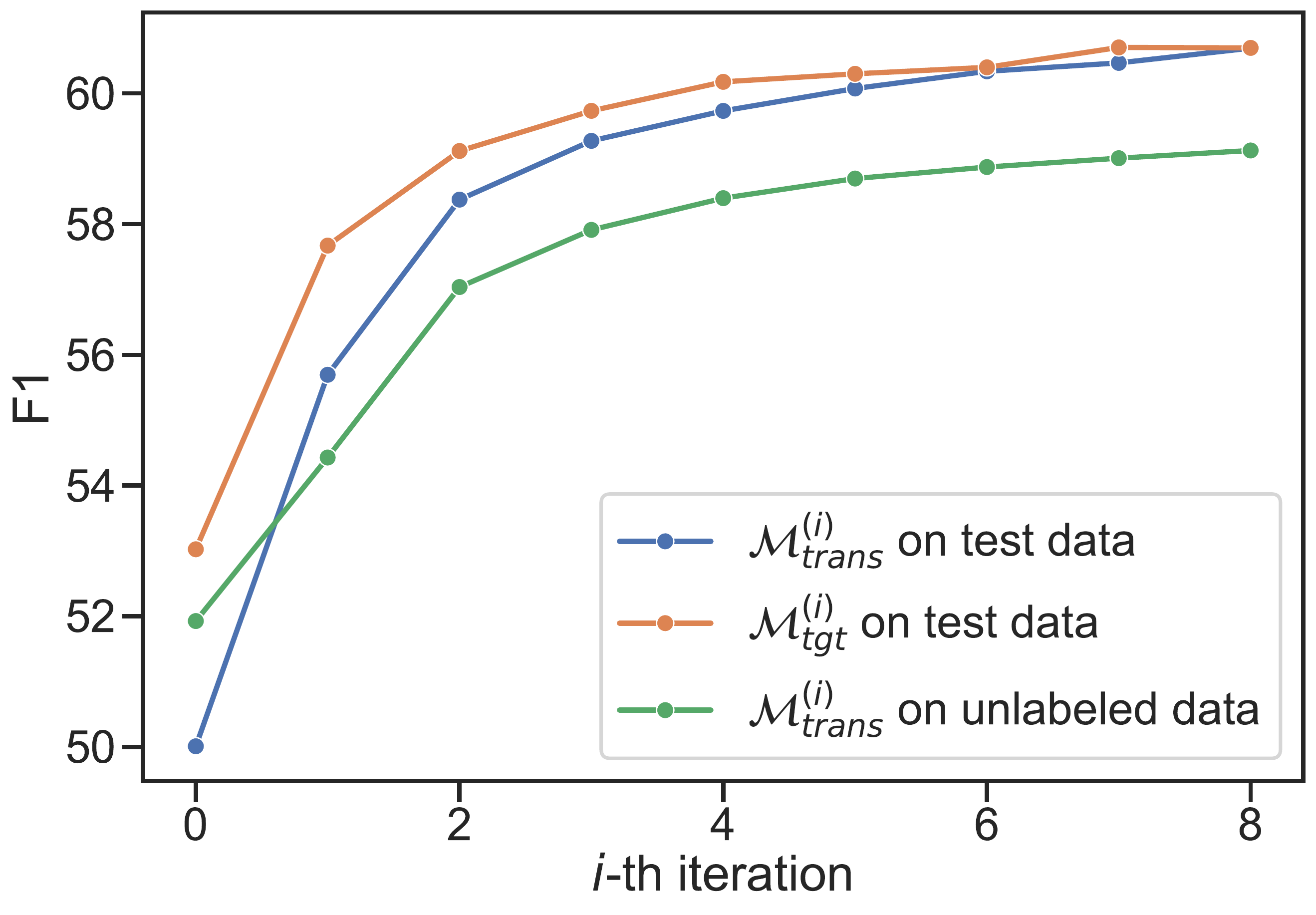}
    \caption{F1 scores of $\mathcal{M}_{trans}^{(i)}$ and $\mathcal{M}_{tgt}^{(i)}$ on the test data and the target-language unlabeled data.
  $\mathcal{M}_{tgt}^0$: model trained on source-language data.
  $\mathcal{M}_{trans}^0$: model trained on original translation data.
    }
    \label{fig:model-iter-zh}
\end{figure}

As shown in Figure~\ref{fig:model-iter-zh}, F1 scores of $\mathcal{M}_{trans}$ and $\mathcal{M}_{tgt}$ consistently improve as iterations go on, and finally converge at the last iteration.
This indicates that both models benefit from the proposed model collaboration scheme.
Two reasons are speculated:
i) An improved $\mathcal{M}_{tgt}$ can provide more accurate labels on the translation data, which further help to improve $\mathcal{M}_{trans}$ via noise-robust learning on such translation data.
For example, at the initial step ($i=0$), the F1 score of the model $\mathcal{M}_{trans}^0$ trained on the original translation labels is 50.0.
With the additional supervision from the collaborator $\mathcal{M}_{tgt}^0$, $\mathcal{M}_{trans}^1$ 
achieves a performance gain of 5.7 F1.
ii) An improved $\mathcal{M}_{trans}$ predicts pseudo labels with higher quality on the target-language unlabeled data, which further benefits the learning of $\mathcal{M}_{tgt}$.
As in Figure~\ref{fig:model-iter-zh}, the quality of pseudo-labeled $\mathcal{D}_{tgt}$ (the green line) grows as $\mathcal{M}_{trans}$ improves.
In this way, both $\mathcal{M}_{trans}$ and $\mathcal{M}_{tgt}$ are providing more and more reliable labels for each other to learn as the iterations progress.

\subsection{Analysis of Instance Collaboration}
This subsection dives into the working mechanism of the instance-collaboration-based denoising.

\begin{figure}[htbp]
    \centering
    \includegraphics[width=0.8\columnwidth]{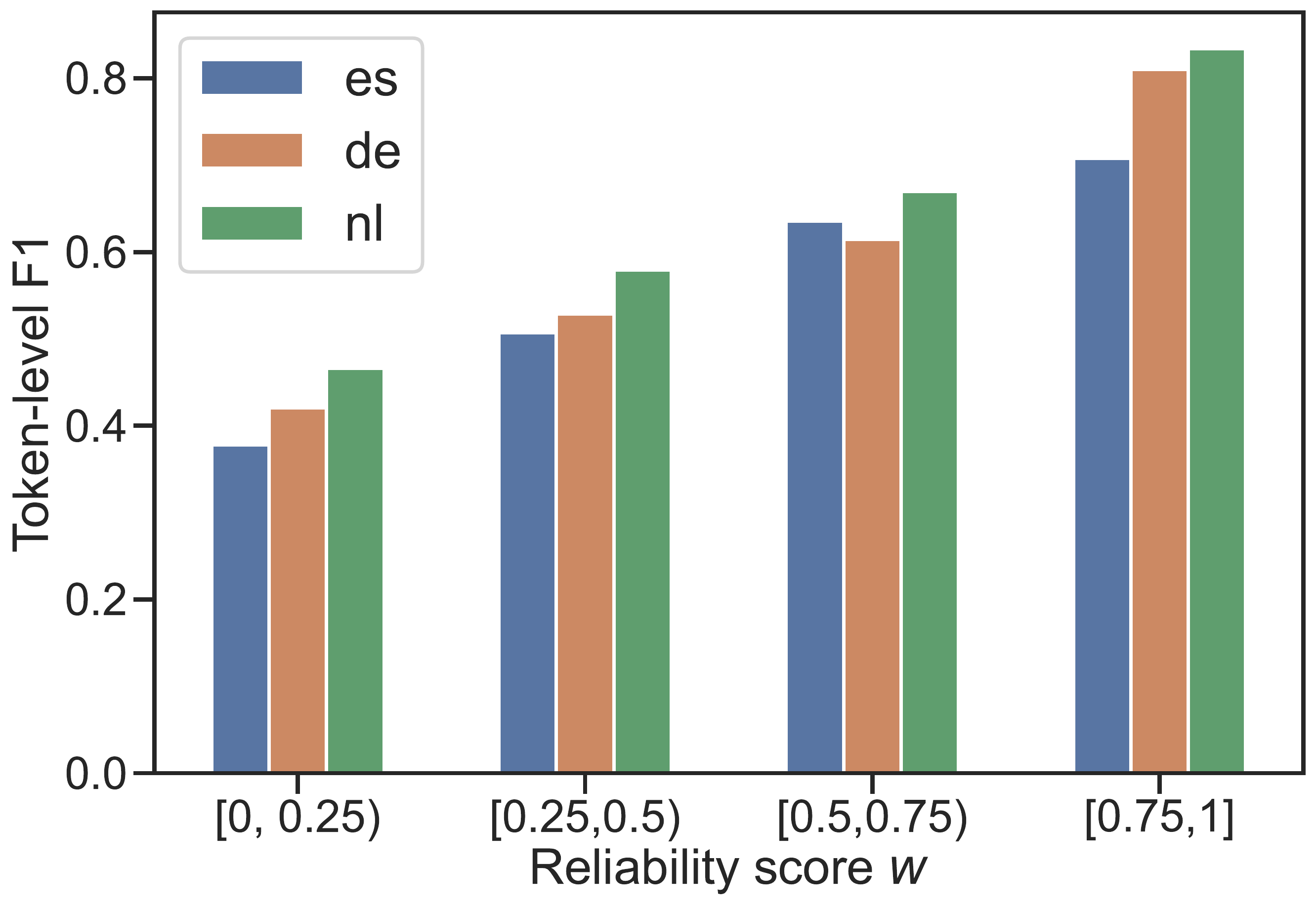}
    \caption{Illustration of the relationship between reliability score and label quality.
    }
    \label{fig:knn-w-conll}
\end{figure}

\begin{figure}[htbp]
    \centering
    \includegraphics[width=0.8\columnwidth]{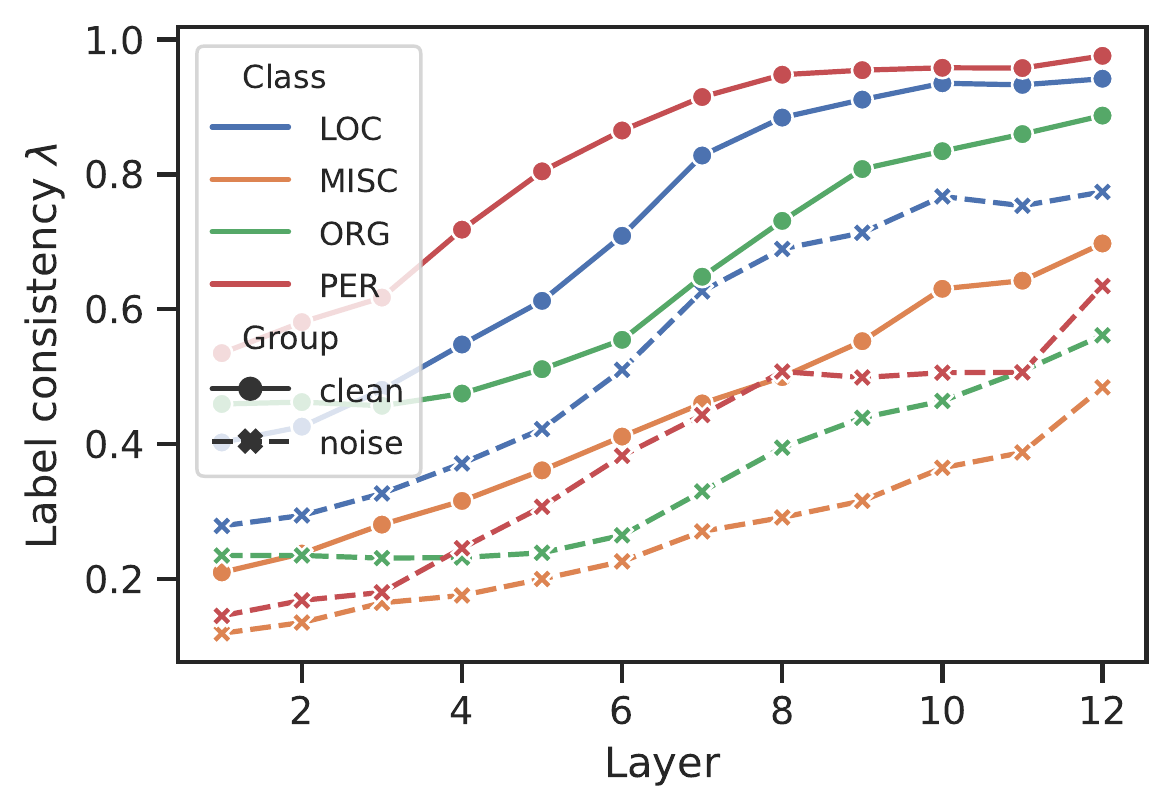}
    \caption{Mean label consistency calculated from different conditions (\ie, different entity types, representation spaces of different layers, clean/noisy tokens) on the German dataset.
    }
    \label{fig:knn-layer}
\end{figure}

\paragraph{Reliability scores v.s. label quality.} 
To study the relationship between reliability score and label quality, 
we partition tokens in the target-language unlabeled data, $x_i\in\mathcal{D}_{tgt}$ into several bins according to their reliability scores $w(p_i^t, x_i)$ calculated via $\mathcal{M}_{trans}^{(1)}$. Then, we compute the token-level F1 over each bin by comparing pseudo labels $\hat{y}_i=\arg \max (p_i^t)$ to the ground-truth ones.
As shown in Figure~\ref{fig:knn-w-conll},
the label quality is proportional to the reliability score, which well demonstrates the effectiveness of our instance-collaboration-based denoising strategy.
\paragraph{Analysis of Label Consistency.}
We also study the characteristics of label consistency \wrt{} different entity types and representation spaces of the memory bank.
Figure \ref{fig:knn-layer} shows the results.
We can draw some in-depth observations as follows.

i) Clean tokens show a larger average consistency than noisy tokens \wrt{} all entity types, demonstrating the effectiveness of our label consistency based denoising strategy again.

ii) Different entity types lead to different distributions of label consistency, which validates the necessity of our design for \textit{class-adaptive} reliability score for weighting as Eq.(\ref{eq:knn}).

iii) Label consistencies calculated with token representations from the upper layers are generally larger than those corresponding to the bottom layers.
Also, the label consistency gap between clean tokens and noisy tokens gets larger from the bottom to the top (\eg, the gap between two orange lines).
This may be attributed to the fact that representations from upper layers are more task-specific~\citep{muller-2021-undermbert}, hence they can better discriminate between noisy and clean tokens.
\paragraph{Choice of K for neighborhood information.}
\begin{figure}[htbp]
    \centering
\includegraphics[width=0.8\columnwidth]{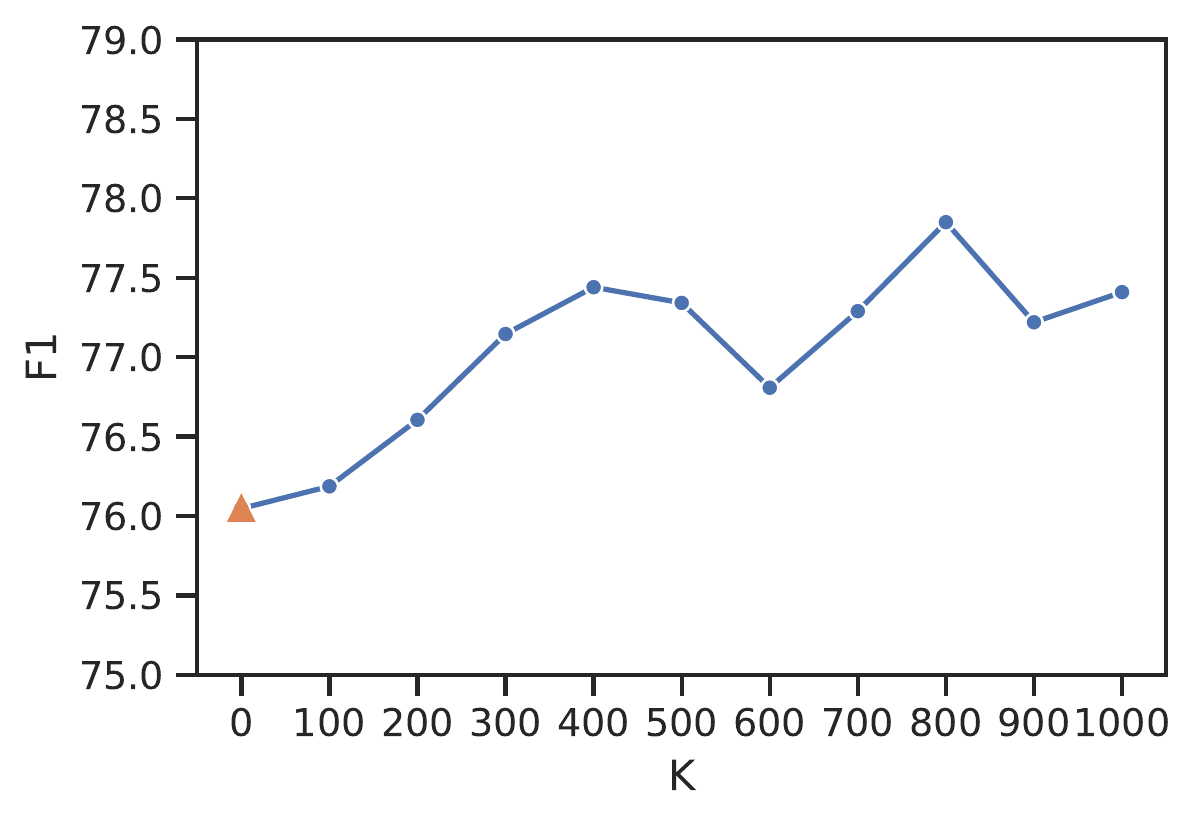}
    \caption{F1 scores of CoLaDa with different $K$ for neighborhood information on German dataset.}
    \label{fig:knn-k}
\end{figure}  
\begin{figure*}[t]
\centering
\includegraphics[width=2\columnwidth]{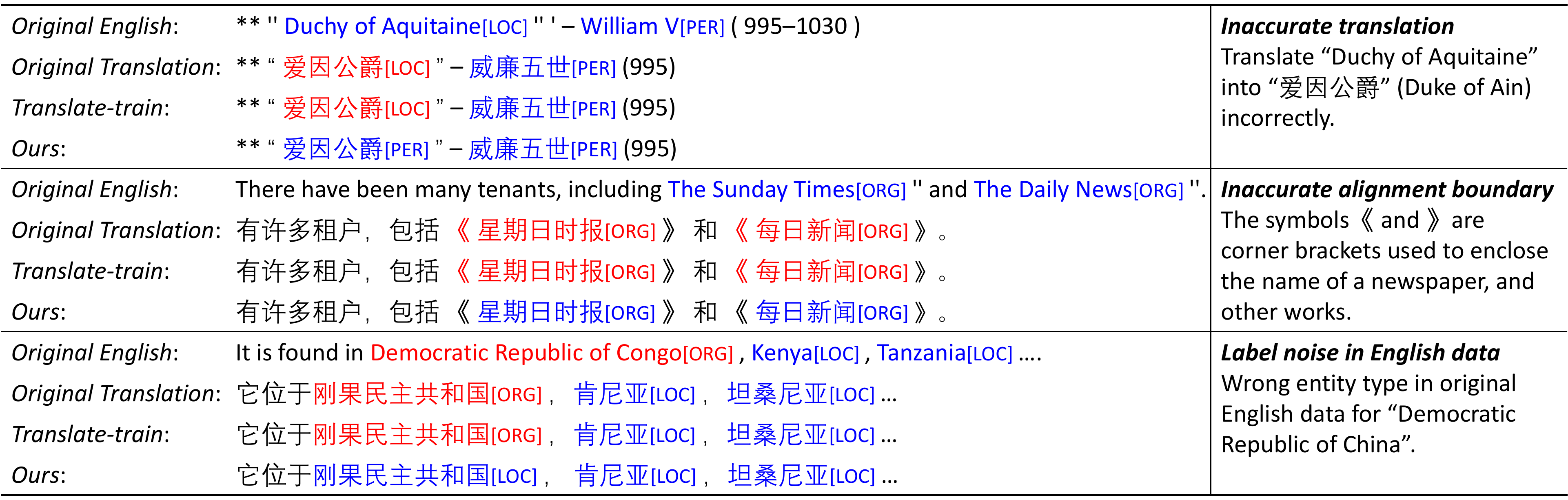}
    \caption{Case study on translation data in Chinese. The \textcolor{blue}{blue} (\textcolor{red}{red}) texts denote the correct (incorrect) entity labels. The \textit{original translation} lines display the translation texts and labels obtained by data projection. \textit{Translate-train} and \textit{Ours} illustrate the predictions from the translate-train method ($\mathcal{M}_{trans}^0$) and our CoLaDa, respectively.}
    \label{fig:case-trans-zh}
\end{figure*}

Figure~\ref{fig:knn-k} shows the performance of CoLaDa using different $K$ in Eq. (\ref{eq:knn-frac}).
Generally speaking, CoLaDa is robust to the choice of K. Any value for $K > 0$ leads to a better performance compared with removing the instance collaboration, \ie{,} K = 0.
A smaller $K$ may lead to a slight performance drop due to limited neighborhood information.

\subsection{Case Study}
To better illustrate the kinds of label noise presented in the data and the capability of CoLaDa to address such noise, we conduct a case study on the Chinese translation data from the WikiAnn English data. As shown in Figure~\ref{fig:case-trans-zh}, there are three typical cases of noisy labels in the translation data: noisy labels induced by inaccurate translations, alignment errors, and annotation errors in the original source-language data.\footnote{Due to the short entity context information in many sentences in WikiAnn, the translation quality of entity mentions with M2M100 is less than satisfactory on the dataset.} Figure~\ref{fig:case-trans-zh} shows that the translate-train model, finetuned on the original translation data, overfits the noisy labels. However, CoLaDa is less affected by such noise and makes correct predictions.

\section{Related Work}

Prior work on cross-lingual NER mainly falls into two major categories: feature-based and data-based transfer.

\paragraph{Feature-based} These methods learn language-independent features so that the model trained on the source language can directly adapt to the target language. 
Earlier work exploits word clusters~\citep{tackstrom-2012-cluster}, gazetteers~\citep{zirikly-2015-cross}, Wikifier features~\citep{tsai-2016-wikifier}, and cross-lingual word embedding~\citep{ni-2017-weakly}, \etc. 
More recently, with the fast growth of multilingual pre-trained language models~\citep{devlin-2019-bert, conneau-2020-xlmr} and their promising results on cross-lingual transfer~\citep{wu-2019-suprise}, lots of studies build upon such pre-trained models and further promote the learning of language-independent features via meta-learning~\citep{wu-2020-meta}, contrastive alignment~\citep{wu-dredze-2020-explicit}, adversarial learning~\citep{keung-2019-adversarial,chen-2021-advpicker}, and by integrating other resources~\citep{fetahu-2022-dygaz}. Despite the great success, they mostly ignore language-specific features, which are especially important when transferring to distant languages~\citep{fu-2023-heter}.

\paragraph{Data-based} These approaches learn language-specific features via automatically labeled target-language data and can be further divided into \textit{translation-based} and \textit{knowledge distillation-based} methods.

Translation-based methods first translate the source-language data to the target language, then perform label projection from the source side to the target side. 
Some prior studies have proposed to use cheap translation such as word-to-word~\citep{xie-2018-word} or phrase-to-phrase~\citep{mayhew-2017-cheap} translation. \citet{jain-2019-entproj} propose an entity projection algorithm to utilize the Google translation system. Recently, \citet{liu-2021-mulda} and \citet{yang-2022-crop} propose to translate sentences with pre-defined markers for label projection. And \citet{ni-2017-weakly} design heuristic rules to select high-quality translation data.
However, both data noise and artifacts~\citep{artetxe-2020-transart} in the translation data still limit the performance of such methods~\citep{garcia-2022-datamodel}. 

Knowledge distillation-based methods train a student model on unlabeled target-language data with the soft labels from a teacher model~\citep{wu-2020-smts}. \citet{li-2022-mtmt} improve the single task based teacher-student learning with entity similarity as an auxiliary task. To mitigate the label noise from the teacher model, \citet{chen-2021-advpicker} propose AdvPicker, which trains a language discriminator to select the less language-dependent unlabeled data for knowledge distillation; \citet{liang-2021-rikd} design a reinforcement learning algorithm to train an instance selector according to features such as model confidence to select reliable pseudo labels iteratively.

While most previous work leverage either translation data or unlabeled data, UniTrans~\citep{wu-2020-unitrans} utilizes the model trained on translation data to perform teacher-student learning on unlabeled data. But it still suffers from the data noise problem. More recently, consistency training~\citep{zheng-2021-xtune, zhou-2022-conner} has also been explored to leverage both unlabeled data and translation data without explicit label annotation. 

To the best of our knowledge, we are the first to propose a unified denoising framework to handle data noise in both translation and unlabeled data \textit{collaboratively} from the model and instance levels for cross-lingual NER.

\section{Conclusion}
To address the problem of label noise in cross-lingual NER, this paper presents CoLaDa, a collaborative label denoising framework. We propose a model-collaboration-based denoising scheme to make two models trained on different data sources to denoise the labels of each other and hence promote each other's learning. We further propose an instance-collaboration-based strategy that collaboratively considers the label consistency among similar tokens in the feature space to re-weight the noisy labels assigned by a teacher model in knowledge distillation. By integrating the instance-collaboration strategy into the model-collaboration denoising scheme, our final framework CoLada achieves  superior performance over prior start-of-the-art methods by benefiting from better handling the data noise.

\section*{Limitations} 
Our framework relies on the availability of translation system and unlabeled data in the target language, which can not be applied to languages without any unlabeled text or translation text. The knowledge distillation step requires a certain amount of unlabeled text, while it may struggle in cases where only few hundreds of unlabeled sentences are available. It would be interesting to combine our label denoising framework with data augmentation techniques in such scenarios.
Besides, the boarder application to other low-resource languages, such as MasakhaNER 2.0~\citep{adelani-2022-masakhaner}, and other cross-lingual sequence labeling tasks are left for exploration in future work.

% \section*{Acknowledgement}
% We thank all reviewers for their valuable suggestions to help us improve this work.

% Entries for the entire Anthology, followed by custom entries
\bibliography{anthology,custom}
\bibliographystyle{acl_natbib}

\appendix
% This is an appendix.
\counterwithin{figure}{section}
\counterwithin{table}{section}

\section{Appendix}
\label{sec:appendix}

\subsection{Dataset Statistics}

Table~\ref{tab:data} reports the dataset statistics for CoNLL and WikiAnn.

\begin{table}[htbp]
    \centering
    \small
    \begin{tabular}{ccccc}
    \toprule
    \textbf{Language} & \textbf{Statistic} & \textbf{Train} & \textbf{Dev} & \textbf{Test} \\
    \midrule
      {English (en)}  & $N_S$ & 14,042 & 3,252 & 3,454 \\
      (CoNLL-2003)  & $N_E$ & 23,499 & 5,942 & 5,648 \\
       \hline
      {German (de)}  & $N_S$ & 12,167 & 2,875 & 3,009 \\
      (CoNLL-2003)  & $N_E$ & 11,851 & 4,833 & 3,673 \\
       \hline
      {Spanish (es)}  & $N_S$ & 8,405 & 1,926 & 1,524 \\
      (CoNLL-2002)  & $N_E$ & 18,798 & 4,351 & 3,558 \\
       \hline
      {Dutch (nl)}  & $N_S$ & 15,836 & 2,895 & 5,202 \\
      (CoNLL-2002)  & $N_E$ & 13,344 & 2,616 & 3,941 \\
       \hline
      {English (en)}  & $N_S$ & 20,000 & 10,000 & 10,000 \\
      (WikiAnn)  & $N_E$ & 27,931 & 14,146 & 13,958 \\
       \hline
      {Arabic (ar)}  & $N_S$ & 20,000 & 10,000 & 10,000 \\
      (WikiAnn)  & $N_E$ & 22,501 & 11,267 & 11,259 \\
       \hline
      {Hindi (hi)}  & $N_S$ & 5,000 & 1,000 & 1,000 \\
      (WikiAnn)  & $N_E$ & 6,124 & 1,226 & 1,228 \\
       \hline
      {Chinese (zh)}  & $N_S$ & 20,000 & 10,000 & 10,000 \\
      (WikiAnn)  & $N_E$ & 24,135 & 12,017 & 12,049 \\
    \bottomrule
    \end{tabular}
    \caption{Dataset statistics. $N_S$: the number of sentences, $N_E$: the number of entities.}
    \label{tab:data}
\end{table}

\subsection{Other Implementation Details}
All experiments are conducted on a Tesla V100 (32GB). The total of trainable parameters ($\mathcal{M}_{trans}$ and $\mathcal{M}_{tgt}$) for our model with mBERT-base-cased as the encoder is 172M and the training time is about 35 mins for one iteration. With XLM-R-large as our base encoder, the total of trainable parameters are 822M and the training takes about 90 mins for one iteration.

\subsection{Baselines}
\label{sec:baseline}
We consider the following start-of-the-art baselines:

\textbf{mBERT}~\citep{wu-2019-suprise} and \textbf{XLM-R}~\citep{conneau-2020-xlmr}  directly train an NER model on the labeled data in the source language, with mBERT and XLM-R as the basic encoder, respectively.

\textbf{BERT-align}~\citep{wu-dredze-2020-explicit} tries to explicitly add word-level contrastive alignment loss to enhance the mBERT representation.

\textbf{AdvCE}~\citep{keung-2019-adversarial} exploits adversarial learning on source- and target-language text to avoid learning language-specific information.

\textbf{AdvPicker}~\citep{chen-2021-advpicker} leverages adversarial learning to learn language-shared features and then selects the less language-specific sentences in target-language unlabeled text for knowledge distillation.

\textbf{MulDA}~\citep{liu-2021-mulda} proposes the labeled sequence translation method for data projection from source-language NER data, a generative model is further applied to augment more diverse examples in the target language.

\textbf{UniTrans}~\citep{wu-2020-unitrans} unifies model- and translation-data-based-transfer via knowledge distillation.

\textbf{TOF}~\citep{zhang-2021-tof} leverages the labeled data for machine reading comprehension task on target language to help the NER task in cross-lingual transfer.

\textbf{TSL}~\citep{wu-2020-smts} proposes knowledge distillation to use unlabeled target-language data for cross-lingual NER.

\textbf{RIKD}~\citep{liang-2021-rikd} proposes a reinforcement learning algorithm to iteratively select reliable pseudo-labels for knowledge distillation.

\textbf{MTMT}~\citep{li-2022-mtmt} proposes multi-task multi-teacher knowledge distillation, which further leverages the entity similarity task.

\textbf{xTune}~\citep{zheng-2021-xtune} leverages unlabeled translation text and other word-level data augmentation techniques for consistency training.

\textbf{ConNER}~\citep{zhou-2022-conner} conducts span-level consistency training on unlabeled target-language data using
translation and further applies dropout-based consistency training on the source-language data.

\end{document}